\documentclass[10pt,twocolumn,letterpaper]{article}

\usepackage{arxiv}
\usepackage{times}
\usepackage{epsfig}
\usepackage{graphicx}
\usepackage{subfigure}
\usepackage{amsmath}
\usepackage{amssymb}
\usepackage{booktabs}
\usepackage{multirow}
\usepackage{algorithm} 
\usepackage{algorithmic}  
\usepackage{enumitem}
\usepackage{authblk}
\makeatletter \renewcommand\AB@affilsepx{\; \protect\Affilfont} \makeatother
\newcommand{\ra}[1]{\renewcommand{\arraystretch}{#1}}
\usepackage[pagebackref=true,breaklinks=true,letterpaper=true,colorlinks,bookmarks=false]{hyperref}

\begin{document}
\title{Generative Hybrid Representations for Activity Forecasting \\ with No-Regret Learning}
\author[1, 2]{Jiaqi Guan\thanks{Work done primarily while Jiaqi Guan and Nicholas Rhinehart were at CMU. Jiaqi Guan's current email: \tt{jiaqi@illinois.edu}}}
\author[1]{Ye Yuan}
\author[1]{Kris M. Kitani}
\newcommand\CoAuthorMark{\footnotemark[\arabic{footnote}]}
\author[1, 3]{Nicholas Rhinehart\protect\CoAuthorMark}
\affil[1]{Carnegie Mellon University}
\affil[2]{University of Illinois Urbana Champaign}
\affil[3]{UC Berkeley \authorcr}
\affil[ ]{\tt\small \{jiaqig, yyuan2, kkitani\}@cs.cmu.edu, nrhinehart@berkeley.edu}

\maketitle
\begin{abstract}

Automatically reasoning about future human behaviors is a difficult problem but has significant practical applications to assistive systems. Part of this difficulty stems from learning systems' inability to represent all kinds of behaviors. Some behaviors, such as motion, are best described with continuous representations, whereas others, such as picking up a cup, are best described with discrete representations. Furthermore, human behavior is generally not fixed: people can change their habits and routines. This suggests these systems must be able to learn and adapt continuously. In this work, we develop an efficient deep generative model to jointly forecast a person's future discrete actions and continuous motions. On a large-scale egocentric dataset, EPIC-KITCHENS, we observe our method generates high-quality and diverse samples while exhibiting better generalization than related generative models. Finally, we propose a variant to continually learn our model from streaming data, observe its practical effectiveness, and theoretically justify its learning efficiency.
\end{abstract}

\vspace{-5mm}
\section{Introduction}
A key requirement for intelligent systems to safely interact with humans is the ability to predict plausible human behaviors. Additionally, they must be able to adapt to variability in behavior over time. However, forecasting a person's behaviors is generally difficult due to the immense set of possible behaviors that humans showcase. This makes it challenging to choose a unified representation for human behavior. Some behaviors are better modeled as continuous representations, for instance, a person's future trajectory. Other behaviors are more succinctly represented discretely, such as interacting with an object. Our goal is to develop an efficient predictive model for joint discrete-continuous spaces, which takes rich sensory information from egocentric videos as input to forecast a person's future behaviors. 

\begin{figure}[t]
\centering
\includegraphics[width=0.99\linewidth]{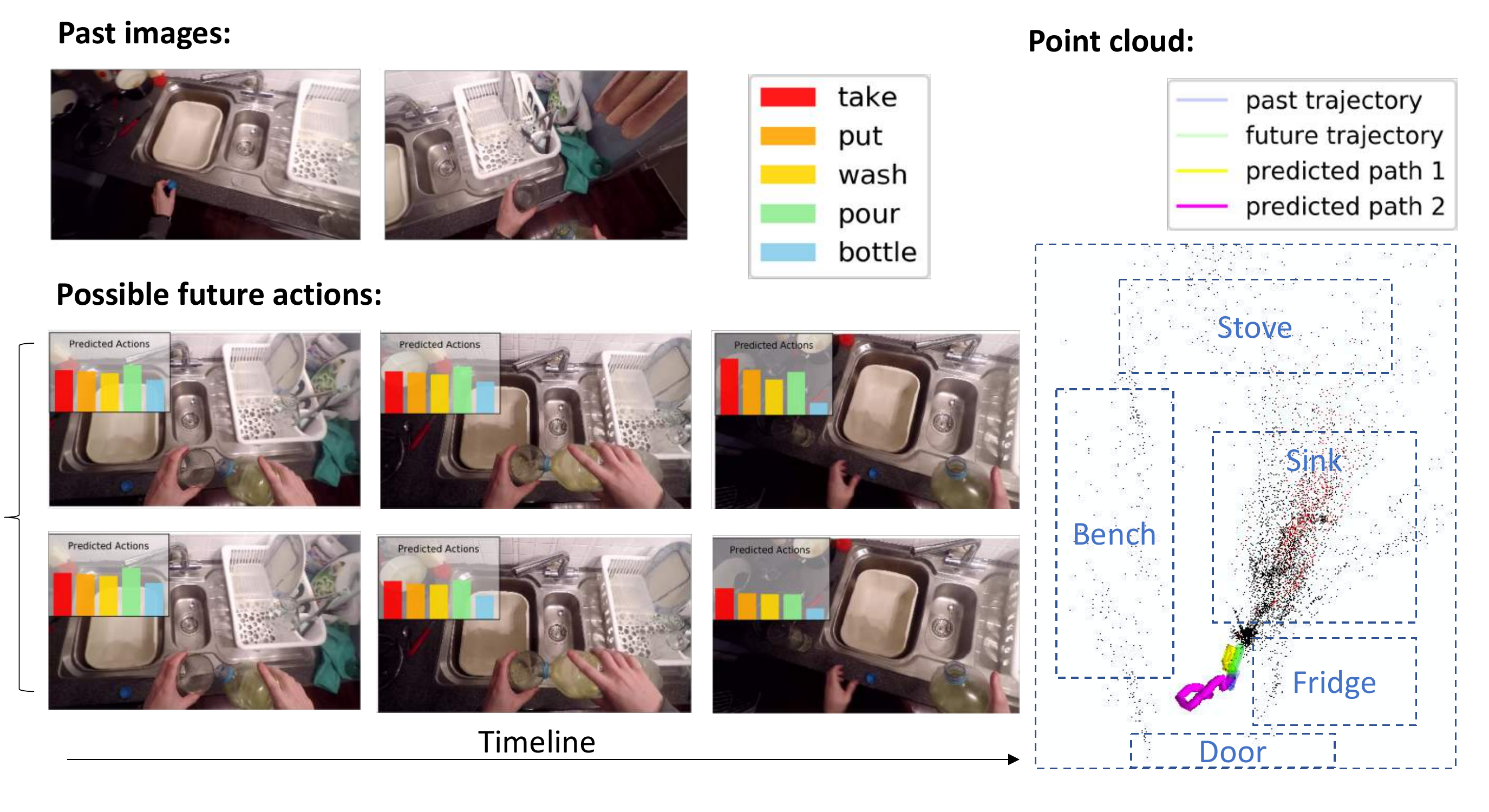}
\caption{\textbf{Generative hybrid activity forecasting.} Our model generates possible future trajectories and actions with past trajectory and images as context. A point cloud is recovered with ORB-SLAM. Histograms show the possibilities of top 5 action classes.}
\label{fig:intro}
\vspace{-5mm}
\end{figure}

For many applications based on a predictive model of future human behavior, it is important that the model is able to characterize the uncertainty of its predictions. A generative model can naturally represent uncertainty and is also well-suited for modeling a hybrid representation of human behavior. Thus, we propose a generative model that can represent the joint distribution of discrete and continuous behaviors by leveraging recent success in generative modeling. Unlike some popular generative models (\emph{e.g.} GANs \cite{goodfellow2014generative} and variational autoencoders~\cite{kingma2013auto}), our method can compute exact likelihoods, which makes it possible to precisely evaluate the model's predictions of future behaviors. It is part of a family of methods known as \emph{invertible generative models} \cite{dinh2016realnvp,grathwohl2018ffjord,kingma2018glow}.   
We learn a generative model of discrete actions by applying the Gumbel-Softmax trick \cite{maddison2016concrete}, and condition this model on continuous samples produced by an invertible generative trajectory model \cite{rhinehart2018r2p2}. We show how we can jointly learn both models efficiently. The results on a large-scale egocentric dataset, EPIC-KITCHENS \cite{damen2018scaling}, demonstrate the advantage of our model in joint trajectory-action forecasting over other generative models and discriminative models. To enable our model to learn optimally from streaming data we employ online learning theories \cite{shalev2012online}. In particular, we apply a modified objective to fine-tune a subset of the model's parameters using a no-regret online learning algorithm. We prove our method's effectiveness theoretically, and observe its online performance matches these theoretical expectations. Example predictions of our method are shown in \ref{fig:intro}.

We present the following contributions: 
\begin{enumerate}
\setlength\itemsep{-3pt}
    \item {\bf Generative hybrid representations:} We propose a generative approach to egocentric forecasting that jointly models trajectory and action distributions. Our experiments on the EPIC-KITCHENS dataset show that our method outperforms both discriminative and generative baselines.
    \item {\bf Exact learning and evaluation:} Our model can compute the probability density function (PDF) exactly and also enables optimization of model sample-based metrics (\emph{e.g.,} reverse cross entropy), which renders learning and inference of people's future trajectory and action more efficient. 
    \item {\bf Theoretically justified no-regret online fine-tuning:} We extend our model to learn online with a simple, yet effective fine-tuning process. We demonstrate that
    it is theoretically efficient, which enables the model to learn from data that arrives continuously and the average \emph{regret} will approach to zero with time elapsing.
\end{enumerate}

\section{Related Work}
We propose a generative model to jointly forecast future trajectories and actions under the first-person vision setting. We begin by discussing work related to our data domain, task, and model. 

\noindent\textbf{First-person vision:}
As wearable cameras become more accessible in our daily lives, a growing body of work is using them for understanding human behaviors \cite{fathi2011understanding, lee2012discovering, ryoo2013first, li2015delving, pirsiavash2012detecting, yuan20183d}. The rich visual information encoded in first-person videos can also be used to predict the subject's attention \cite{li2015delving, yuan2019ego} and their interactions with the environment.

\noindent\textbf{Trajectory Forecasting:}
\emph{Third-person} trajectory forecasting has enjoyed significant research attention recently. The approach in \cite{lee2016predicting} predicts future trajectories of wide-receivers from surveillance video. A large body of work has also used surveillance video to predict future pedestrian trajectories~\cite{Xie2013InferringM, ma2017forecasting, ballan2016knowledge, kitani2012activity}. Deterministic trajectory modeling has been used for vehicle \cite{jain2016recurrent} and pedestrian \cite{alahi2016social, robicquet2016learning, yagi2018future} trajectory prediction. Due to the uncertain nature of future trajectories, modeling stochasticity can help explain multiple plausible trajectories with the same initial context. Several approaches have tried to forecast distributions over trajectories \cite{lee2017desire, galceran2015multipolicy}. \cite{rhinehart2018r2p2} proposed a generative approach to model vehicle trajectories. A relative small amount of work has investigated trajectory forecasting from \emph{first-person} videos. \cite{soo2016egocentric} predicts the future trajectories of the camera wearer by constructing an EgoRetinal map. 

These approaches employed continuous representations in the batch learning setting, while our model uses both discrete and continuous representations in both the batch and online learning settings.

\noindent\textbf{Action Forecasting:}
Classification-based approaches \cite{hoai2014max, lan2014hierarchical, ryoo2011human, ryoo2015robot} are popular in action forecasting. Many activities are best represented as categories. \cite{gao2017red} proposed an encoder-decoder LSTM model to predict future actions. Other work has also tried to forecast more generalized action such as gaze \cite{zhang2017deep}, user-object interactions \cite{furnari2017next} and the position of hands and objects \cite{fan2017forecasting}. In \cite{rhinehart2017first}, online inverse reinforcement learning (IRL) is used to model a person's goals and future trajectories. IRL has also been applied to forecast the behaviors of 
robots \cite{ratliff2006maximum}, taxis \cite{ziebart2008maximum}, and pedestrians \cite{kitani2012activity}. Some work has investigated non-discriminative modeling of future actions. \cite{vondrick2016anticipating} devised a deep multi-modal regressor to allow multiple future predictions. \cite{fan2017forecasting} uses a variational autoencoder (VAE) to model the distribution of possible future actions. Whereas prior activity forecasting approaches reason about actions only, our method reasons jointly about actions and trajectories. 

\noindent\textbf{Generative Models:}
Deep generative models, \emph{e.g.}  \cite{goodfellow2014generative, kingma2013auto}, are a powerful unsupervised modeling approach. To enable efficient learning of deep generative models of categorical distributions, \cite{maddison2016concrete} proposed the Gumbel-Softmax trick to backpropagate gradients through these distributions. There has been work that uses generative models to address the uncertainty in both trajectory \cite{lee2017desire, galceran2015multipolicy, rhinehart2018r2p2,yuan2019diverse,WengYuan2020,yuan2020dlow} and action forecasting \cite{fan2017forecasting, vondrick2016anticipating}. Unlike the prior approaches, our method jointly generates the future trajectories and actions.

\noindent\textbf{Online Learning:}
The field of online learning studies how to learn effectively from streaming data \cite{shalev2012online}, but these approaches are rarely used in computer vision problems. In \cite{rhinehart2017first}, online inverse reinforcement learning is performed with visual data. In contrast, our approach is based on imitation learning without reward modeling. In~\cite{ross2011reduction, ross2014reinforcement}, interactive imitation learning is framed as a online learning problem. Our approach, while a form of imitation learning, is not interactive. It observes expert behavior (human behaviors) and makes predictions that the human does not interact with.
\section{Generative Hybrid Activity Forecasting}

\begin{figure*}[t]
\begin{center}
   \includegraphics[width=0.85\linewidth]{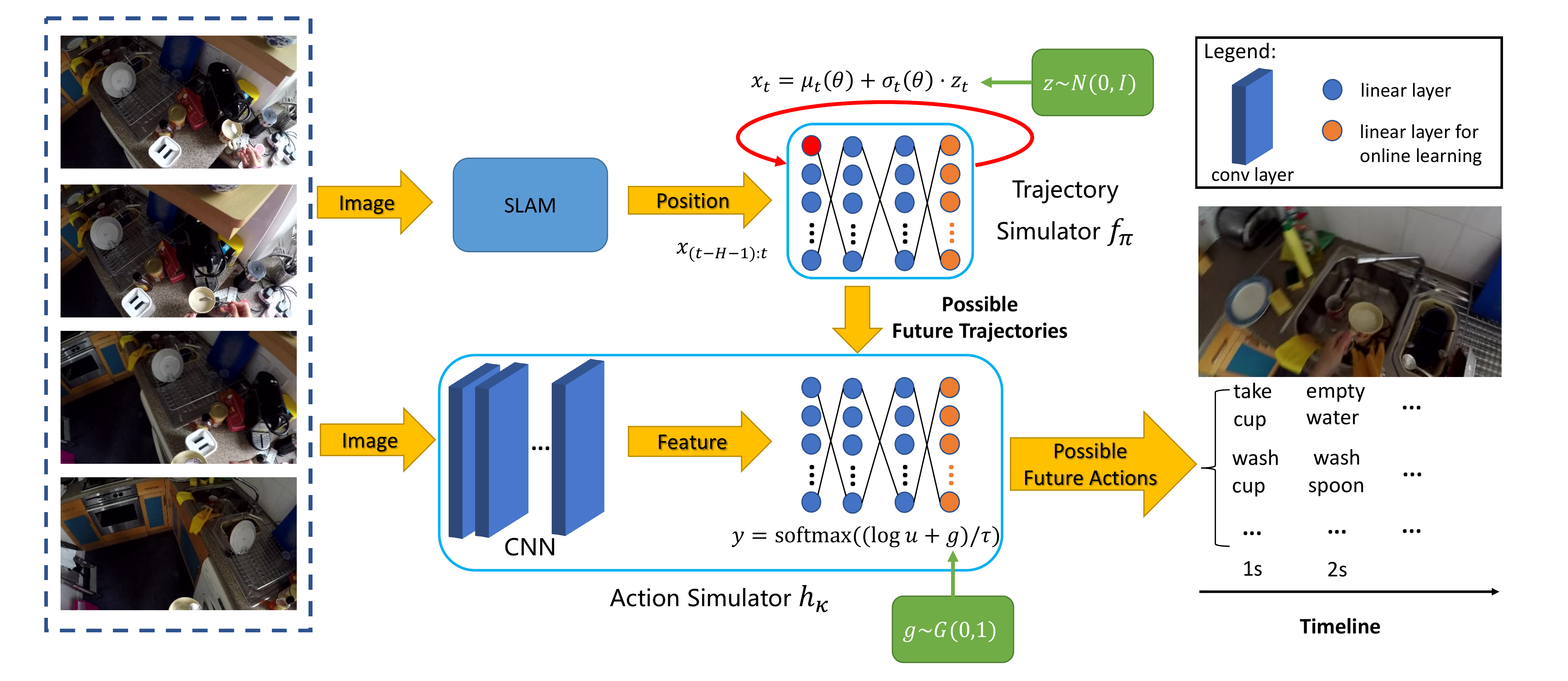}
\end{center}
\vspace{-3mm}
\caption{\textbf{Our proposed model.} ORB-SLAM \cite{murTRO2015} is used to extract positions from videos. Trajectory simulator $f_\pi$ takes past positions and noise sequence from Gaussian distribution to generate future trajectory. Action simulator $h_\kappa$ takes past images and positions as well noise sequence from Gumbel distribution to produce future actions.}
\vspace{-3mm}
\label{fig:model}
\end{figure*}

\subsection{Problem Formulation}

Our goal is to model the true joint distribution $p(x, a|\phi)$ of a person's future trajectory $x\in \mathbb{R}^{T\times 3}$ in 3D and actions $a\in \{0, 1\}^{T\times C_a\times 2}$ from egocentric videos with a learned joint distribution $q(x, a|\phi)$, where $\phi$ is the context information, $T$ is the forecasting horizon, and $C_a$ is the number of action classes (with each class modeled with 2 values using a one-hot encoding). The context information $\phi$ includes past egocentric video frames $V_{-P:0}$ and positions $x_{-P:0}$, where $P$ is the observation horizon. 

As $x$ and $a$ use different representations (continuous {vs.} discrete), we further factorize the joint distribution by conditioning $a$ on $x$ i.e. $q(x, a|\phi) = q(x|\phi)q(a|x,\phi)$. Learning this model of future behavior via divergence minimization is akin to \emph{imitation learning} \cite{ghasemipour2019divergence,ke2019imitation}. We use \emph{one-step} policies $\pi$ for generating trajectory $x$ and $\kappa$ for generating actions $a$, and samples from $q_\pi(x|\phi)$ and $q_\kappa(a|x,\phi)$ can be obtained by repeatedly sampling $T$ times from $\pi$ and $\kappa$. These policies parameterize each generative model. Our training data is a set of episodes denoted $\{(x, a, \phi)_n\}_{n=1}^N$, which are samples from the (unknown) data distribution of the person's behavior $p(x,a|\phi)$. We use this data to train the policies $\pi$ and $\kappa$, thereby learning $q(x,a|\phi)$.

\subsection{Complementary Cross Entropy Loss}

A desired feature of forecasting models is to generate both \emph{diverse} and \emph{precise} predictions. Following~\cite{rhinehart2018r2p2}, we construct a complementary cross-entropy loss to train our trajectory-action distribution $q(x, a|\phi)$:
\begin{small}
\begin{equation} 
\label{eq:sym_loss}
    \mathcal{L} = \underbrace{\mathbb{E}_{(x, a)\sim p}-\log q(x, a|\phi)}_{H\left(p, q\right)} + \beta \underbrace{\mathbb{E}_{(x, a)\sim q}-\log \tilde{p}(x, a|\phi)}_{H\left(q, \tilde{p}\right)}\,,
\end{equation}
\end{small}
where $\tilde{p}$ is an approximation to the data distribution $p$, which we will discuss it in detail in Sec \ref{sec:prior_dist}.  $\beta$ is a weighting factor.
The forward cross entropy term $H(p,q)$ encourages the distribution $q$ to cover all modes of $p$ and thus increases sample diversity. The reverse cross entropy term $H(q, \Tilde{p})$ penalizes samples far from the data distribution $\tilde{p}$ to improve sample quality. The joint use of 
them promotes both diversity and quality of samples. We use $\beta$ to control the trade-off between diversity and precision.

With the factorization $q(x,a|\phi) = q_\pi(x|\phi)q_\kappa(a|x,\phi)$, the forward and reverse cross entropies can be rewritten as
\begin{small}
\begin{equation}
\label{eq:cond_loss}
\begin{aligned}
    H\left(p, q\right) &= \underbrace{-\mathbb{E}_{x\sim p}\log q_\pi\left(x|\phi\right)}_{H(p, q_\pi)} \underbrace{- \mathbb{E}_{(x,a)\sim p}\log q_\kappa\left(a|x,\phi\right)}_{H(p, q_\kappa)}\, , \\
    H\left(q, \Tilde{p}\right) &= \underbrace{-\mathbb{E}_{x\sim q_\pi}\log \Tilde{p}\left(x|\phi\right)}_{H(q_\pi, \tilde{p})} \underbrace{-\mathbb{E}_{x\sim q_\pi, a\sim q_\kappa}\log \Tilde{p}\left(a|x,\phi\right)}_{H(q_\kappa, \tilde{p})}\, .
\end{aligned}
\end{equation}
\end{small}

This decomposition disentangles the cross entropies for trajectory and actions, allowing us to learn the policy $\pi$ and $\kappa$ separately. The optimization of $H(p, q)$ requires us to compute $q$ and the optimization of $H(q, \tilde{p})$ requires us to sample from $q$. Different from GANs \cite{goodfellow2014generative} (likelihood-free learning) and VAEs \cite{kingma2013auto} (optimize the evidence lower bound), we propose an \emph{invertible generative model}, which enables us to both compute the likelihood of $q(x, a|\phi)$ exactly and generate samples from $q(x, a|\phi)$. The model details will be illustrated in Sec \ref{sec:pos_ce}, \ref{sec:act_ce} and \ref{sec:policy_modeling}.  

\subsection{Trajectory Cross Entropy}
\label{sec:pos_ce}
We employ an invertible trajectory generative model by constructing a differentiable, invertible function $f_\pi(z;\phi): \mathbb{R}^{T\times3} \rightarrow \mathbb{R}^{T\times3}$. This function maps a noise sequence $z = [z_1, \dots, z_T]$ from a Gaussian distribution $\mathcal{N}(0, I_{3 \times 3})$ and the scene context $\phi$ to a trajectory $x = [x_1, \dots, x_T]$. $f_\pi$ is implemented by a $\theta$-parametrized per-step policy $\pi$. At each time step $t$, $\pi$ takes in a per-step context $\psi_t$, containing past positions $x_{t-P:t-1}$, and outputs the mean $\mu_t$ and an invertible covariance matrix $\sigma_t$, and simulate the current position $x_t$ with noise $z_t$: $x_t \triangleq \mu_t\left(\psi_t;\theta\right) + \sigma_t\left(\psi_t;\theta\right)z_t \,.$ Since $\sigma_t$ is invertible, $\pi$ defines a bijection between $z_t$ and $x_t$, and $f_\pi$ defines a bijection between $x$ and $z$.

$q_\pi$ then follows from the change-of-variables formula for multivariate integration \cite{rezende2015variational,dinh2016realnvp,grathwohl2018ffjord,kingma2018glow}:
\begin{small}
\begin{equation}
    q_\pi\left(x|\phi\right) = \mathcal{N}\left(f_\pi^{-1}(x;\phi)\right)|\text{det} J_{f_\pi}\left(f_\pi^{-1}(x;\phi)\right)|^{-1}\,,
\end{equation}
\end{small}
where $J_{f_\pi}(f_\pi^{-1}(x;\phi))$ is the Jacobian of $f_\pi$ evaluated at $f_\pi^{-1}(x;\phi)$. Thus, the forward cross entropy can be rewritten as
\begin{small}
\begin{equation}
    \label{eq:traj_fce} H(p, q_\pi) = -\mathbb{E}_{x\sim p} \log \frac{\mathcal{N}\left(f_\pi^{-1}(x;\phi)\right)}{|\text{det} J_{f_\pi}\left(f_\pi^{-1}(x;\phi)\right)|} \,.
\end{equation}
\end{small}

The reparameterization also greatly simplifies the differentiation of $H(q_\pi, \tilde{p})$ w.r.t. policy $\pi$. Instead of sampling from $q_\pi$, we can sample from $\mathcal{N}$ and rewrite the reverse cross entropy as Eq.\eqref{eq:traj_rce}. $z$ is the source of uncertainty for generating diverse samples.
\begin{small}
\begin{equation}
    \label{eq:traj_rce}
    H(q_\pi, \tilde{p}) = -\mathbb{E}_{z\sim \mathcal{N}}\log\Tilde{p}\left(f_\pi(z;\phi)|\phi\right) \,.
\end{equation}
\end{small}

\vspace{-3mm}
\subsection{Action Cross Entropy}
\label{sec:act_ce}
For the action forecasting, at each step $t$ each single action class $c$ is represented as $a_{t, c} \in \{0, 1\}^2$ which is a one-hot vector indicating whether this action happens $([0, 1])$ or not $([1, 0])$. Since actions are discrete variables, we use Gumbel-Softmax distributions \cite{jang2016categorical} to reparameterize actions. We build a simulator $h_\kappa(g;\phi): \mathbb{R}^{T \times C_a \times 2} \rightarrow \{0, 1\}^{T \times C_a \times 2}$, which maps noise sequences $g$ sampled from Gumbel distribution $\mathcal{G}(0, 1)$ to actions $a$. The noise sequence $g$, as a key part of the Gumbel-Softmax reparameterization -- a continuous, differentiable approximation to Gumbel-Max, provides an efficient way to draw samples from a categorical distribution. 

The per-step action forecasting context $\chi_t$ consists of past images $V_{-P:0}$ and past positions $x_{t-P:t-1}$. The per-step policy $\kappa$ outputs action probabilities $u_t$ with $\chi_t$, and simulate the current action $a_t$ with noise $g_t$: 
\begin{small}
\begin{equation*}
    a_{t,c,i} \triangleq \frac{\exp\left((\log(u_{t,c, i}(\chi_t;\theta)) + g_{t,c,i}) / \tau\right)}{\sum_{j=1}^2 \exp\left((\log(u_{t,c,j}(\chi_t;\theta) + g_{t,c,j}) / \tau\right)} \,,
\end{equation*}
\end{small}
where $i \in \{1,2\}$, $c \in \{1,\dots,C_a\}$, and $t\in\{1,\dots,T\}$, $\tau$ is the temperature of Gumbel-Softmax distribution.

According to the probability density function of the Gumbel-Softmax distribution \cite{jang2016categorical}, the action forward cross entropy can be rewritten as
\begin{small}
\begin{equation}
\label{eq:act_fce}
\begin{aligned}
    &H\left(p, q_\kappa\right) =\\
    &-\mathbb{E}_{(x, a) \sim p}\sum_{t,c} \log \tau\! \left(\sum_{i=1}^2\frac{u_{t,c, i}\left(\chi_t\right)}{a_{t,c,i}^\tau}\right)^{\!-\!2}\prod_{i=1}^2\left(\frac{u_{t,c, i}\left(\chi_t\right)}{a_{t,c,i}^{\tau+1}}\right),
\end{aligned}
\end{equation}
\end{small}
\hspace{-1mm}
For the reverse cross entropy, using Gumbel-Softmax reparameterization, it can be rewritten as
\begin{small}
\begin{equation}
\label{eq:act_rce}
    H\left(q_\kappa, \Tilde{p}\right) =  -\mathbb{E}_{g\sim \mathcal{G}} \sum_{t,c,i}\log\Tilde{p}(a_{t,c,i}|x,\phi) \,.
\end{equation}
\end{small}

The overall procedure of training the batch model is shown in Algorithm \ref{algo:batch}.

\begin{algorithm}
	\caption{\small{Offline Generative Hybrid Activity Forecasting}}
	\begin{algorithmic}[1]
	  \label{algo:batch}
	  \REQUIRE{Training dataset $\{(x, a, \phi)_n\}_{n=1}^N$; Batch size $B$; Trajectory simulator $f_\pi$; Action simulator $h_\kappa$}
	  \STATE Randomly initialize $f_\pi$ and $h_\kappa$ with parameter $\theta$
	  \REPEAT
	  \FOR{each mini-batch examples $(x, a, \phi)_{i:i+B}$}
	    \STATE Calculate $H(p, q_\pi)$ with Eq. \eqref{eq:traj_fce} \eqref{eq:act_fce}
	    \STATE Sample $z\sim \mathcal{N}$; Generate trajectory $\hat x = f_\pi(z; \phi)$
	    \STATE Calculate $H(q_\pi, \Tilde{p})$ with Eq. \eqref{eq:traj_rce}
	    \STATE Sample $g\sim \mathcal{G}$; Generate actions $\hat a = h_\kappa(g; \phi)$
	    \STATE Calculate $H(q_\kappa, \Tilde{p})$ with Eq. \eqref{eq:act_rce}
	    \STATE Update $\theta$ by optimizing Eq.\eqref{eq:sym_loss} 
	  \ENDFOR
	  \UNTIL{$\theta$ converge}
	  \RETURN $\theta$ as $\hat \theta$
	\end{algorithmic}
\end{algorithm}

\begin{algorithm}
	\caption{\small{Online Generative Hybrid Activity Forecasting}}
	\begin{algorithmic}[1]
	  \label{algo:online}
	  \REQUIRE{Trajectory simulator $f_\pi$; Action simulator $h_\kappa$; Pre-trained weights $\hat \theta$}
	  \STATE Initialize $f_\pi, h_\kappa$ with $\hat \theta$
	  \STATE Fix all parameters except the linear layer $\theta_0$ at the end
	  \FOR{each new example}
	    \STATE $[x, y, z] \xleftarrow{}$ slam.track()
	    \STATE Calculate $H(p, q)$ with Eq. \eqref{eq:traj_fce} \eqref{eq:act_fce}
	    \STATE Sample $z\sim \mathcal{N}$; Generate trajectories $\hat x = f_\pi(z; \phi)$
	    \STATE Calculate $H(q, \Tilde{p})$ with Eq. \eqref{eq:online_rce}
	    \STATE Finetune $\theta_0$ by optimizing Eq. \eqref{eq:online_loss} with SGD
	  \ENDFOR
	\end{algorithmic}
\end{algorithm}

\subsection{Policy Modeling}
\label{sec:policy_modeling}

\noindent\textbf{Trajectory Modeling.}
For the trajectory policy $\pi$, we use a recurrent neural network (RNN) with gated recurrent units \cite{cho2014learning} that maps context $\psi_t$ to $\hat{\mu}_t$ and $S_t$.
We use the matrix exponential~\cite{najfeld1995derivatives} to ensure the positive definiteness of $\sigma_t$: $\sigma_t = \text{expm} \left(S_t + S_t^T\right)$.
The network architecture is shown in Figure \ref{fig:model}. We provide more architectural details in Appendix \ref{apsec:arch}.

\vspace{1mm}
\noindent\textbf{Action Modeling.}
Our action policy $\kappa$ maps context $\chi_t$ to action probabilities $u_t$, and is based on the idea of Temporal Segment Networks \cite{wang2016temporal} with a ResNet-50 \cite{he2016deep} backbone. The past images $V_{-P:0}$ we observe are divided into $K$ segments and an image is selected randomly from each segment. These images are passed through a ResNet independently to get the class scores. Another fully-connected layer is built on top of the ResNet to fuse these class scores to yield segmental consensus, which serves as a useful feature in our action forecasting. In the meanwhile, the past trajectory $x_{t-P:t-1}$ also includes useful information about what kind of actions people may perform. Thus, we add a MLP which takes the segmental consensus and the past trajectory as inputs to generate the action probabilities ~$u_t$.

\subsection{Prior Distribution Approximation}
\label{sec:prior_dist}
It is challenging to evaluate $H(q_\pi, p)$ without the PDF of $p$ (here, the density function of future behavior). We propose a simple approach to estimate it using the training data. For trajectory $H(q_\pi, p)$, we build $\Tilde{p}$ as a sequence of unimodal normal distributions with ground-truth trajectory $\tilde{x}$ as means, \emph{i.e.}, $\tilde{p}(x|\phi) = N(\cdot|\tilde{x}; \sigma I)$. In fact, this is identical to adding a mean squared distance penalty between the predicted trajectories and expert trajectories. For action $H(q_\kappa, p)$, we first assume that if an action occurs at time $t$, then the same action has a higher probability happening at time steps closer to $t$. Based on this assumption, we can also view each action happening at $t$ as a unimodal normal distribution in the time dimension. If the action spans several time steps, we take the max of the distributions induces by different $t$. As a result, we obtain the approximate action prior distribution $\tilde{p}(a|x,\phi)$. Note that this action prior does not actually depend on the trajectory $x$, this is partly due to the difficulty of defining a conditioned prior distribution. On the other hand, our reverse cross entropy can be seen as a regularization of trajectories and action, and the independent version can achieve this.

\subsection{Online No-regret Learning}

To apply the proposed framework to an online scenario where the policies are learned over time, we would like to ensure that the learning process is guaranteed to converge to the performance of the strongest model. We can evaluate the relative convergence properties of an online learning algorithm through regret analysis. To leverage known proofs of no-regret learning, one should ensure that the model and loss function being used is convex. To this end, we pretrain the network and fix parameters of nonlinear layers. We slightly adjust the trajectory reverse cross entropy as Eq. \eqref{eq:online_rce} and perform online gradient descent on the loss function in Eq.~\eqref{eq:online_loss} by fine-tuning the parameters of the last linear layer. The regret is computed with respect to a model family, and the model family we consider is one of pre-trained representations that are fine-tuned to adapt to online performance. The detailed online learning parameterization is explained in Appendix \ref{apsec:online}.
\begin{small}
\begin{equation}
\label{eq:online_rce}
    H\left(q_\pi, \Tilde{p}\right)^\text{adj} = -\mathbb{E}_{x_{1:t\!-\!1} \sim p, x_{t:T} \sim q_\pi} \log\Tilde{p}\left(x|\phi\right) \,,
\end{equation}
\begin{equation}
\label{eq:online_loss}
    L_\text{online} = H\left(p, q_\pi\right) + H\left(q_\pi, \Tilde{p}\right)^\text{adj} + H\left(p, q_\pi\right)\,.
\end{equation}
\end{small}

In general, the regret $R_T$ of an online algorithm is defined as:
$R_T = \sum_{t=1}^T l_t\left(\xi_t;\theta_t\right) - \min_{\theta^*}\sum_{t=1}^T l_t\left(\xi_t;\theta^*\right)$, where $\xi_t$ and $l_t$ is the input and the loss at time step $t$ separately. We can prove our forward cross entropy loss is convex with respect to the parameters of the finetuned linear layer. If we further constrain the parameter's norm $\|\theta\|_2 \leq B$ and the gradient's norm $\|\nabla_\theta\|_2 \leq L$, then the regret of our online algorithm is bounded \cite{shalev2012online} as: $R_T \leq BL\sqrt{2T} \,.$

Since the bound is sub-linear in $T$, the average regret $R_T/T$ approaches zero as $T$ grows, so it is a no-regret algorithm. The overall online learning procedure is shown in Algorithm \ref{algo:online}. The detailed proof of the no-regret property is given in Appendix \ref{apsec:online} and the empirical results are shown in the experiments.

\section{Experiments}
We evaluate our models and baselines on the EPIC-KITCHEN \cite{damen2018scaling} dataset. In this section, we first describe the dataset and related data processing steps. We then introduce the baselines that we use to compare our model with and the metrics to evaluate the performance of trajectory forecasting and action forecasting. 
In the experiments, we perform both batch and online experiments with the goal to validate the following hypotheses: (1) Since the trajectory-action joint model make actions conditioned on positions, the extra position information should help achieve better action forecasting performance than separately trained model. (2) The reverse cross entropy terms for trajectory and actions in our loss function should help improve sample quality. (3) The ability of evaluating the exact PDF of the trajectory and action distribution should help our model achieve lower cross entropies and higher sample quality than other generative methods that do not optimize the exact PDF such as CVAE. 
(4) The generative model should have the ability to generate samples with higher quality than discriminative models since it considers the multi-modal nature of future behavior and can generate multiple reasonable samples during the evaluation, while discriminative models can not.
(5) We want to show from an empirical perspective that our online learning method is effective and no-regret.

\subsection{Data Description}
We evaluate our method on the EPIC-KITCHENS dataset \cite{damen2018scaling}. First, we use ORB-SLAM \cite{murTRO2015} to extract the person's 3D positions from the egocentric videos. For each video, we start to collect positions when the pose-graph is stable and no global bundle adjustment is performed. We also scale positions with the first 30-second results by assuming that the person's activity range in each video is similar to alleviate the scale ambiguity caused by the initialization of ORB-SLAM. Then, we extract examples with successive 7-second interval. Those discontinuous examples (such as when tracking gets lost) are dropped out. In each 7-second example, we use the past 2 seconds as context to predict the future trajectory and actions in the next 5 seconds. We down-sample original data to 5 fps for position, 2 fps for images, and 1 fps for actions. Thus, the context we use to train the model contains 10 past positions and 4 past images. We filter actions to guarantee that each action occurs at least 50 times and drop videos which includes less than 5 examples. Finally, we use 4455 examples in total, which come from 135 videos. The number of action classes is 122 with 39 verbs and 83 nouns. 
Since the annotations of the test set are not available, we randomly split the original training videos to training, validation, and test with the proportion of 0.7, 0.1, 0.2. At the same time, we ensure each action occurs in both training set and test set and the examples in different sets come from different videos.  

We predict verbs and nouns separately instead of predicting the pairs of them, which is different from the setting in \cite{damen2018scaling}. This is because first the combination of verbs and nouns would create too many action classes and each class would have few samples; second, there are often multiple actions taking place at the same time in the dataset, which leads to our multi-label classification formulation.

\subsection{Baselines and Metrics}

\paragraph{Baselines}
The baselines we use include two generative models and a discriminative model:
\begin{itemize}[leftmargin=*]
\setlength\itemsep{-3pt}
\item \textbf{Direct Cross Entropy (DCE):} a generative model that uses a sequence of Gaussian to model the trajectory distribution, and a sequence of Bernoulli distributions conditioned on the trajectory to model the action distribution.
\item \textbf{Conditional Variational Autoencoder (CVAE):} an auto-regressive variant VAE-based generative model. We use the Gumbel-Softmax to model the action distribution.
\item \textbf{Mixed Regression and Multi-label Classification (MRMC):} a discrimintative model trained by  minimizing the mean squared error of trajectories and the binary cross entropy of actions.
\end{itemize}
For all baseline models, we follow the same network structure as our model to process past positions and images context. Detailed info can be found in Appendix \ref{apsec:arch}.

\vspace{1mm}
\noindent\textbf{Metrics}
We use the following metrics to comprehensively evaluate our method and other baselines:
\begin{itemize}[leftmargin=*]
\setlength\itemsep{-3pt}
\item \textbf{Forward Cross entropy:} for trajectory and action forecasting, we use their corresponding forward cross entropies $H(p, q_\pi)$ and $H(p, q_\kappa)$ to evaluate how well the policy mimics the behaviors of the expert.
\item \textbf{minMSD and meanMSD:} for trajectory forecasting, we also include two common sample-based metrics often used in generative models -- minMSD and meanMSD~\cite{lee2017desire, walker2016uncertain, gupta2018social, rhinehart2018r2p2}. minMSD computes the smallest distance from $K$ samples to the ground-truth $x$: $\min_k \| \hat{x}^k - x \|^2$. Thus, minMSD evaluates the quality of the best sample. In contrast, meanMSD evaluates the overall quality of all $K$ samples via $\frac{1}{K}\sum_{k=1}^K \|\hat{x}_k\ - x\|^2$. The combined use of these two metrics evaluates the quality of generated trajectories comprehensively. We sample 12 trajectories for each example.
For discriminative models, we directly report the regression results as minMSD and meanMSD.
\item \textbf{Precision, Recall and F-1 score:} for action forecasting, since the action space is large and we need to forecast actions in 5 seconds per example, the exact matching accuracy is not be a good metric. Instead, we calculate the example-based precision and recall as \cite{zhang2014review}.
One special case is that if there is no ground-truth action or predicted action happening at some time step, the denominator will be zero. If this happens, the precision and recall is 1 only if $tp = fp = fn = 0$, where $tp$, $fp$, $fn$ is the number of true positives, false positives, and false negatives, otherwise the precision and recall is 0. 
To consider both precision and recall, we also calculate F-1 score as $F_1 = \frac{2 \times \text{precision} \times \text{recall}}{\text{precision} + \text{recall}}$. As action distribution is conditioned on the forecasted trajectory, we first sample 12 trajectories, and for each trajectory we sample the action (for each action class, the action happens if its logit is greater than 0.5) and average the metrics across the trajectories. For discriminative models, we directly report the multi-label classification results.
\end{itemize}

\begin{table*}[htbp]
\centering
\ra{1.2}
\footnotesize
\resizebox{.8\textwidth}{!}{
\begin{tabular}{@{}llllcllll@{}}\toprule
\multirow{2}{*}{Method} & \multicolumn{3}{c}{Trajectory Forecasting} & \phantom{abc}& \multicolumn{4}{c}{Action Forecasting} \\
\cmidrule{2-4} \cmidrule{6-9}
& $H(p, q_\pi)$ $(\downarrow)$& minMSD$(\downarrow)$ & meanMSD$(\downarrow)$ && $H(p, q_\kappa)$ $(\downarrow)$& Precision$(\uparrow)$ & Recall$(\uparrow)$ & $F_1$ $(\uparrow)$\\ \midrule
(a) MRMC  & -   & 0.392  & \textbf{0.392}  && -  & 40.64 & 32.12 & 35.88 \\
(b) DCE & -26.93 & 0.539 $\pm$ 0.010 & 1.870 $\pm$ 0.094 && -40.22 & 11.04 $\pm$ 3.11 & \textbf{39.31} $\pm$ 2.10 & 17.24 $\pm$ 2.49 \\
(c) CVAE & $\leq$ -129.78 & 0.319 $\pm$ 0.008 & 1.394 $\pm$ 0.085 && $\leq$ -135.21  & 38.48 $\pm$ 0.09 & 31.03 $\pm$ 0.04 & 34.38 $\pm$ 0.06 \\
(d) Ours$\,$(S)-F & -288.26 & 0.304 $\pm$ 0.017 & 1.553 $\pm$ 0.077 && -192.48 & 39.06 & 29.97 & 33.92 \\
(e) Ours$\,$(S) & -275.81 & \textbf{0.286} $\pm$ 0.007 & 0.915 $\pm$ 0.088 && -192.31 & 39.97 & 29.80 & 34.14 \\
(f) Ours~(J)-F & \textbf{-298.92} & 0.291 $\pm$ 0.017 & 1.446 $\pm$ 0.087 && -192.53 & 42.89 $\pm$ 0.33 & 32.50 $\pm$ 0.29 & 36.98 $\pm$ 0.30\\
(g) Ours~(J) & -298.47 & 0.293 $\pm$ 0.004 & 0.971 $\pm$ 0.078 && \textbf{-192.57} & \textbf{44.10} $\pm$ 0.11 & 33.39 $\pm$ 0.07 & \textbf{37.90} $\pm$ 0.09 \\
\bottomrule
\end{tabular}
}
\vspace{1em}
\caption{\textbf{Batch results on the EPIC-KITCHENS dataset.} For sample-based metrics, mean $\pm$ std is reported. MRMC: Mix Regression and Multi-label Classification (discriminative model). DCE: Direct Cross Entropy (generative model). CVAE: Conditional Variational Autoenconder (generative model). For our model, $S$ denotes separate training of trajectory policy and action policy. $J$ denotes joint training. $F$ denotes the model is trained with forward cross entropy only. 
$(\downarrow)$/($\uparrow$) denotes a metric for which lower/higher scores are better.}
\label{tab:results}
\end{table*}

\begin{figure*}[htbp]
\centering
\includegraphics[width=0.85\linewidth]{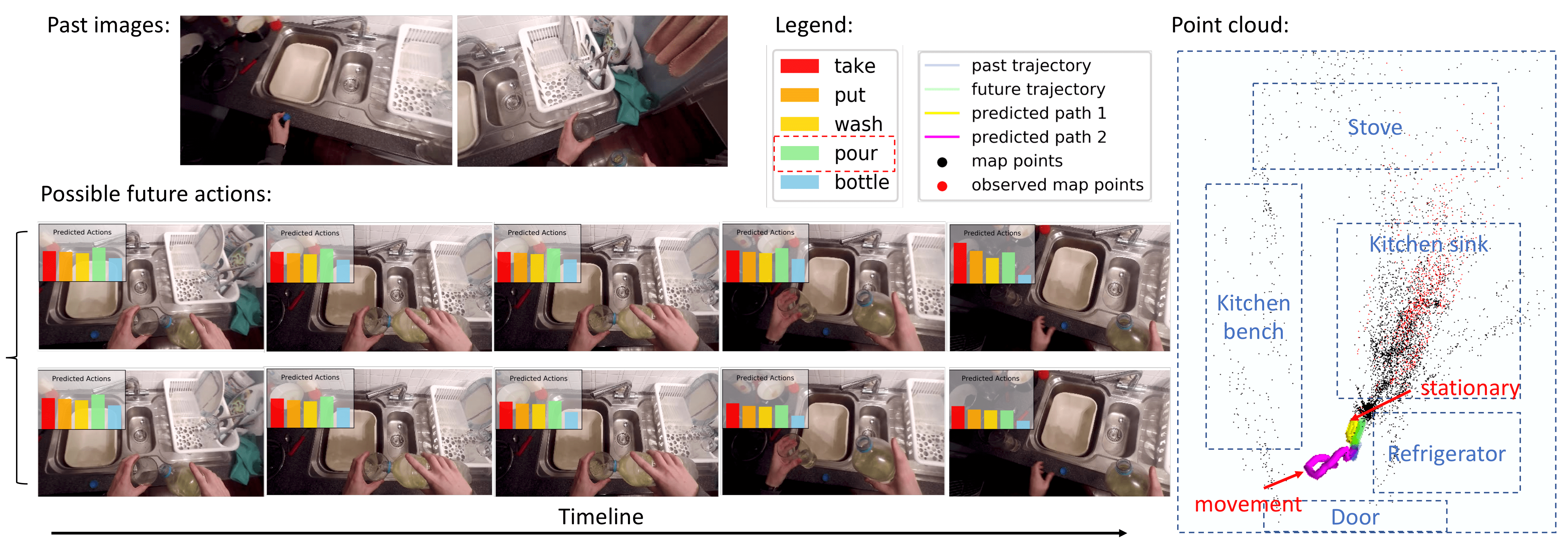}
\vspace{2mm}

\includegraphics[width=0.85\linewidth]{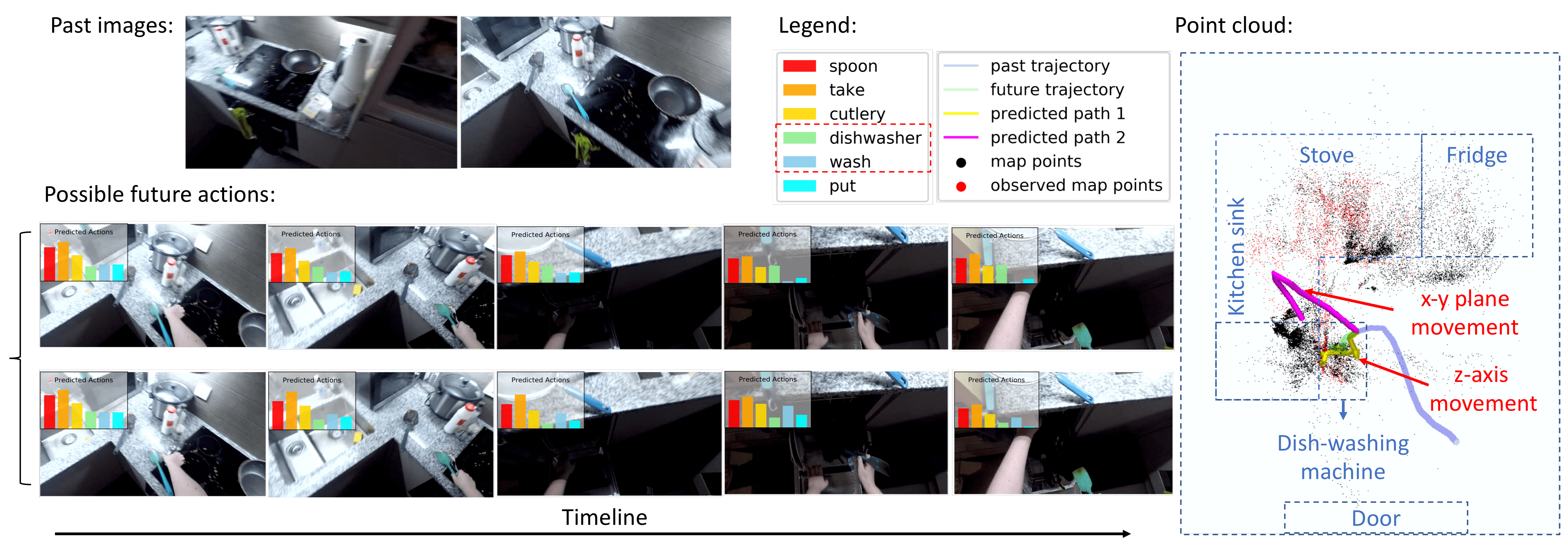}
\caption{\textbf{Forecasting results visualization.} 
Visualization of two examples.
It shows how the forecasted trajectory influences the action distribution. In each example, the left top shows observed images
, the left bottom shows action distributions corresponding to two forecasted sample trajectories, and the right shows the point cloud of the scene and the forecasted trajectories (Red/Black points: Observed/Unobserved map points).}
\vspace{-5mm}
\label{fig:vis}
\end{figure*}

\begin{table*}\centering
\ra{1.2}
\footnotesize
\resizebox{.85\textwidth}{!}{
\begin{tabular}{@{}lllllcllll@{}}\toprule
\multirow{2}{*}{Experiment} & \multirow{2}{*}{Method} & \multicolumn{3}{c}{Trajectory Forecasting} & \phantom{abc}& \multicolumn{4}{c}{Action Forecasting} \\
\cmidrule{3-5} \cmidrule{7-10}
&& $H(p, q_\pi)$ $(\downarrow)$ & minMSD$(\downarrow)$ & meanMSD$(\downarrow)$ && $H(p, q_\kappa)$ $(\downarrow)$ & Precision $(\uparrow)$& Recall $(\uparrow)$& $F_1$ $(\uparrow)$\\ \midrule
\multirow{2}{*}{(i) Train$\to$Test} & Pre-online & -298.47 & 0.293 $\pm$ 0.004 & 0.971 $\pm$ 0.078 && -192.57 & 44.10 $\pm$ 0.11 & \textbf{33.39} $\pm$ 0.07 & 37.90 $\pm$ 0.09\\
& Online & \textbf{-299.66} & \textbf{0.283} $\pm$ 0.004 & \textbf{0.963} $\pm$ 0.063 && \textbf{-192.59} & \textbf{45.27} $\pm$ 0.10 & 32.90 $\pm$ 0.07 & \textbf{38.11} $\pm$ 0.10\\ \midrule
\multirow{2}{*}{(ii) Test$\to$Train} & Pre-online & -204.23 & 0.280 $\pm$ 0.005 &0.560 $\pm$ 0.080 && -181.80 & 20.70 $\pm$ 0.03 & 20.28 $\pm$ 0.02 & 20.49 $\pm$ 0.02\\
& Online & \textbf{-220.38} & \textbf{0.230} $\pm$ 0.004 & \textbf{0.497} $\pm$ 0.091 && \textbf{-184.89} & \textbf{22.76} $\pm$ 0.05 & \textbf{22.05} $\pm$ 0.04 & \textbf{22.40} $\pm$ 0.05\\
\bottomrule
\end{tabular}
}
\vspace{1em}
\caption{\textbf{Online learning results.} \emph{Pre-online} denotes the results on the streaming data before online learning. \emph{Online} denotes the results on the streaming data across online learning. Experiment \emph{$A\to B$} means we pretrain the model on the data set $A$ and perform online learning on the data set $B$.  
$(\downarrow)$/($\uparrow$) denotes a metric for which lower/higher scores are better.
}
\vspace{-3mm}
\label{tab:online_results}
\end{table*}

\subsection{Batch Forecasting Results}
Our full model is a joint forecasting model which makes actions conditioned on the trajectory, and it is trained using the complementary loss function in Eq.~\eqref{eq:sym_loss}.
To test whether the joint modeling of trajectory and action distribution help improve forecasting performance, we also train a trajectory forecasting model and an action forecasting model separately. We also evaluate a variant of our method by using only the forward cross entropy for both action and trajectory. The results are summarized in Table \ref{tab:results}. 

\begin{figure}[htbp]
\centering
\includegraphics[width=0.99\linewidth]{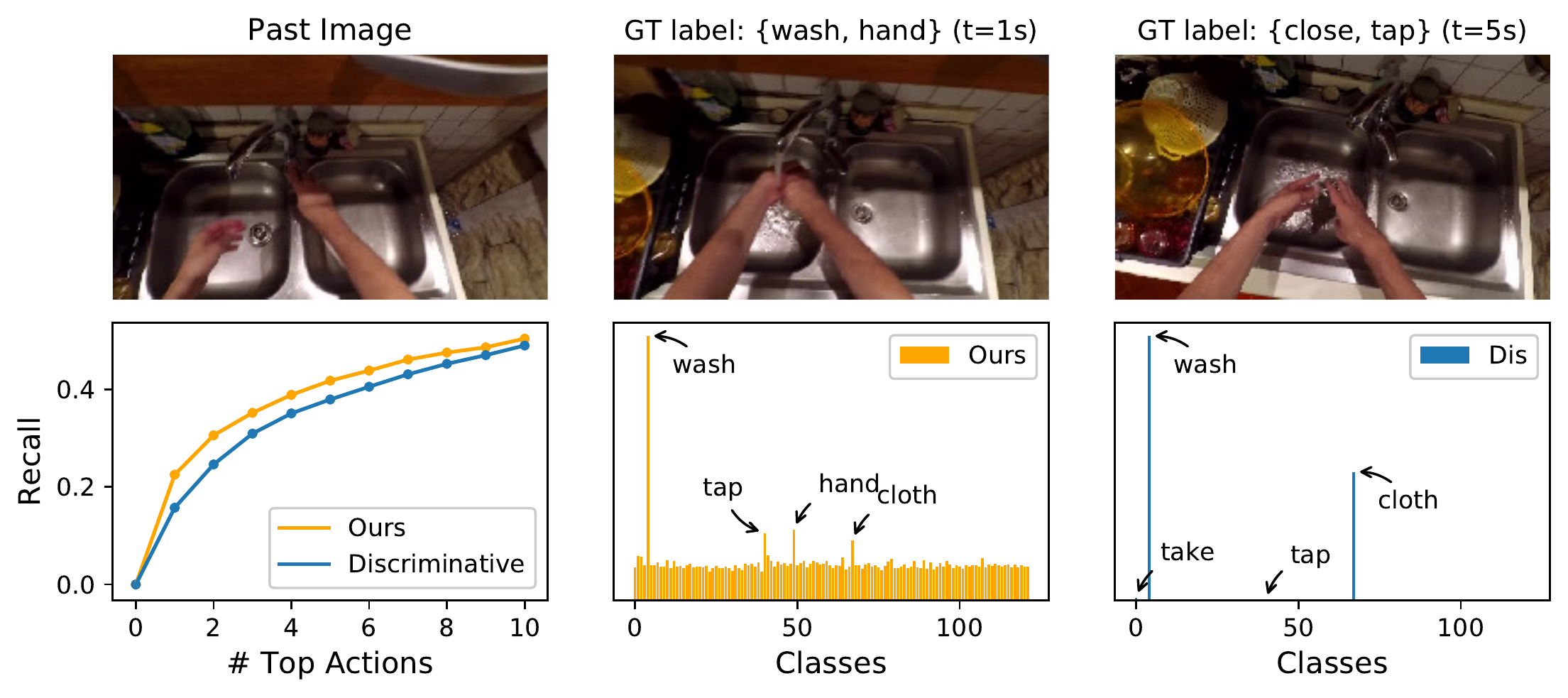}
\vspace{2mm}

\caption{\textbf{Top-K visualization.} The first row is an example of uncertain future behavior. The bottom left plot shows the recall value of our model and the discriminative model if we force the model to output actions with top $K$ probabilities ($K$ is from 1 to 10). The other two plots in the bottom row show the action class probabilities of our model and the discriminative model separately. Our model shows better performance in handling uncertainty.}
\vspace{-5mm}
\label{fig:topK_vis}
\end{figure}

First, we can see that our joint forecasting model (g) outperforms separately trained models (e) in action forecasting metrics (cross entropy, precision, recall, and F1-score), so our factorization -- conditioning actions on the trajectory indeed helps. Hypothesis (1) is supported.
Comparing (e)(g) to (f)(d), we can see the quality of both trajectory samples and actions samples are better after using the reverse cross entropy, which justifies its use in the loss function and also demonstrates the effectiveness of our designed prior data distribution. Hypothesis (2) is supported.
Furthermore, our methods outperforms other generative baselines (b)(c) in terms of most metrics, especially forward cross entropy. This is due to the fact that our method has more modeling power than DCE, and can evaluate the exact PDF of trajectory and action distribution instead of optimizing the variational lowerbound like CVAE does. Our model does not outperform \emph{MRMC} in the meanMSD metric and \emph{DCE} in the Recall metric, but we note that: 1. The \emph{MRMC} model can not conduct sampling, so it leads to lower meanMSD than all of other generative models; 2. The \emph{DCE} model actually cannot generate good enough examples, which is indicated by the low precision and low F1 score, even if it has high recall; 3. All baselines make actions conditioned on positions, so it is fair to compare \emph{Ours(J)} with baselines, which shows better performance except for two aforementioned special cases. Hypothesis (3) is supported. 
Finally, our method also performs better than the discriminative baseline \emph{MRMC}, because it fails to model the multi-modal nature of the future behavior. Fig.~\ref{fig:topK_vis} illustrate this point further. We can see that our model continuously outperforms the discriminative model in terms of recall when we force the model output actions with top $K$ ($K$ is from 1 to 10) probabilities. The visualization example shows an environment with uncertainty. Given past information, we are actually not sure which actions (\emph{wash hand}, \emph{close tap}, \emph{take cloth} or \emph{dry hand}) will happen. Our model assigns relatively high confidence on these probable future actions but the discriminative model only focuses on two actions -- \emph{wash} and \emph{cloth}. Thus, hypothesis (4) is also supported.

Fig.~\ref{fig:vis} shows visualization results of two examples. For each example, we show two sampled trajectories and their corresponding action distribution. In all these two examples, the forecasted trajectory influences the action distribution in a meaningful way. In the first example, the person is going to pour the drink. We can see that the person moves less in the first forecasted trajectory than the second one. As a result, the first trajectory has a higher probability for pouring because people tend to stay still when they are pouring something. In the second example, the person is going to take the cutlery and insert them into the dishwasher. In the first predicted trajectory, the person's position changes a lot along the z-axis, and the model predicts that the person is more likely to put the cutlery into the dishwasher. In contrast, the positions change a lot in the ground plane (xy-plane) in the second forecasted trajectory, and the model predicts that the person is more likely to wash something as it requires more horizontal movements.

\subsection{Online Forecasting Results}
\begin{figure}[htbp]
\centering
\subfigure
{
\label{fig:traj_regret}
\includegraphics[width=0.4\linewidth]{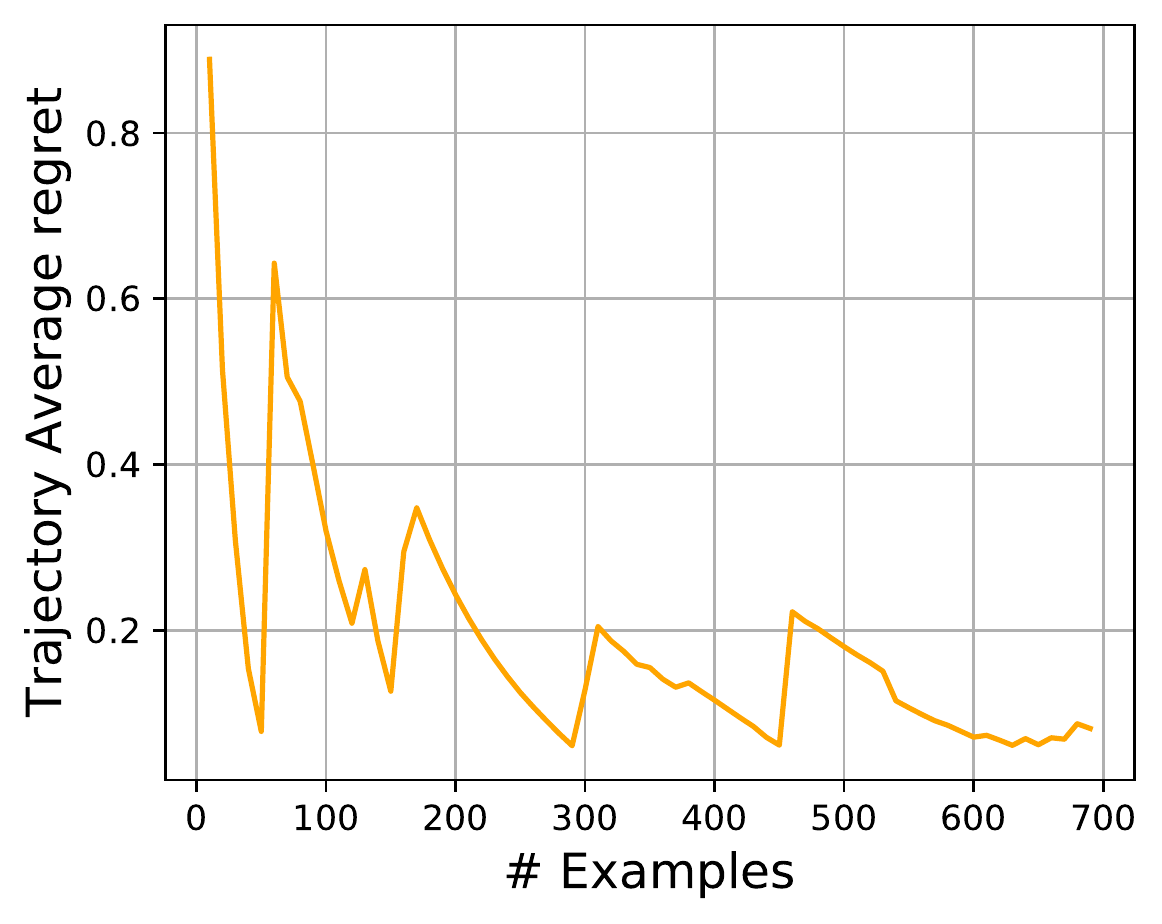} 
}
\subfigure
{
\label{fig:act_regret}
\includegraphics[width=0.41\linewidth]{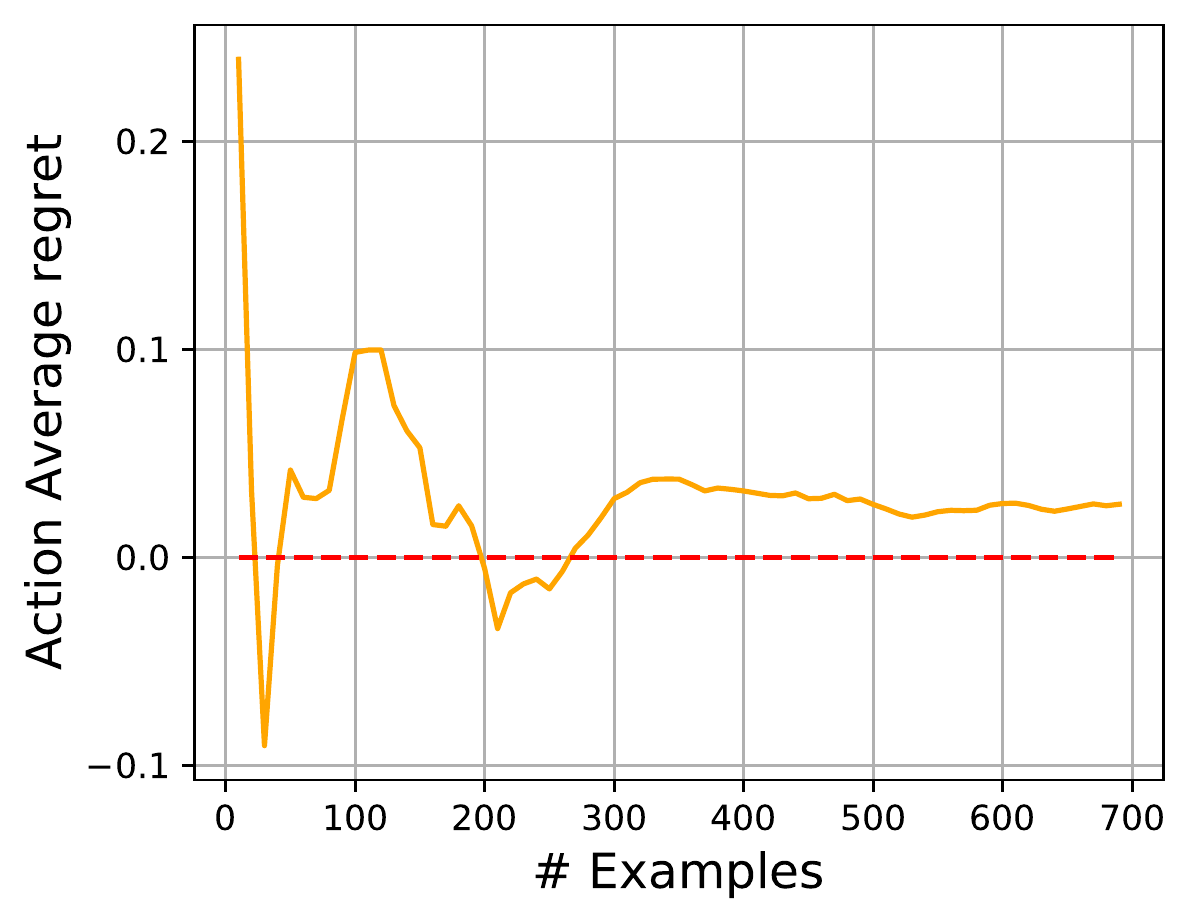}
}
\caption{\textbf{Average regret curve.} We compare our online learning model with the hindsight model to calculate the average regret. The average regret of both trajectory forecasting (left) and action forecasting (right) show convergence towards zero, which supports the claim that our online learning method is no-regret empirically.}
\label{fig:regret}
\vspace{-3mm}
\end{figure}

We conduct two online learning experiments to verify the effectiveness of our model to learn from streaming data. We pretrain the model on the training set and perform online learning on the test set in (i), and inversely in (ii). In both experiments, we only finetune additional linear layers during online learning. Pre-online learning and online learning results are shown in Table~\ref{tab:online_results}. It can be seen that in both experiments, the model obtained after online learning outperforms the original model which shows the effectiveness of our online learning algorithm. Additionally, comparing (ii) with (i), we can also see that with more data observed, the relative improvement from online learning will be more significant.  We also analyze the regret of our model. We train the online models and corresponding hindsight models using Eq.~\eqref{eq:online_loss}. The average regret curve of the forward experiment is shown in Fig.~\ref{fig:regret}. We can see that the average regret curve converges to zero as more examples are observed, which proves that our model is no-regret. Hypothesis (5) is also supported. The theoretical analysis of no-regret can be found in Appendix \ref{apsec:online}. 

\section{Conclusion}
We proposed a novel generative model to represent hybrid continuous and discrete state for first-person activity forecasting. We model discrete actions conditioned on continuous trajectories and learn a deep generative model by minimizing a symmetric cross entropy loss. Our model can generate both precise and diverse future trajectories and actions based on observed past images and positions. The results on EPIC-KITCHENS dataset shows our method outperforms related generative models and discriminative models. Our model can also be easily adapted to no-regret online learning, which creates more application possibilities in complex real-world scenarios. A possible future work is the united representation of continuous and discrete variables with the help of discrete normalizing flow models, instead of factorizing the joint distribution to make actions conditioned on trajectories. 

\noindent\textbf{Acknowledgement.}
This work was sponsored in part by Center for Machine Learning and Health PhD Fellowship, and JST CREST (JPMJCR14E1). 

{\small
\bibliographystyle{ieee}
\bibliography{reference}
}

\clearpage
\onecolumn
\appendix
\section{Proof of Regret Bound}
\label{apsec:online}
According to Equation 2.5 of \cite{shalev2012online}, if a loss function $f$ parameterized by $\theta$ is convex and $\|\theta\|_2 \leq B$ , then the regret of online gradient descent $R_T$ at the $T$-th time step is bounded as follows:
\begin{equation}
    R_T \leq \frac{1}{2\lambda} \|\theta_t\|_2^2 + \lambda \sum_{t=1}^T \|\nabla_{\theta_t}\|_2^2 \, .
\end{equation}

This regret bound depends on the norm of the sub-gradients produced by the algorithm, and thus is not satisfactory. To derive a more concrete bound, according to Corollary 2.7 of \cite{shalev2012online}, if $f_t$ is $L_t$-Lipschitz with respect to $\|\cdot\|_2$, and let $L$ be such that $\frac{1}{T}\sum_{t=1}^T L_t^2 \leq L^2$, then we have:
\begin{equation}
    R_T \leq \frac{1}{2\lambda} \|\theta\|_2^2 + \lambda TL^2 \, .
\end{equation}
In particular, if $\lambda = \frac{B}{L\sqrt{2T}}$, then
\begin{equation}
    R_T \leq BL\sqrt{2T} \, .
\end{equation}
It is obvious that the average regret $\frac{R_T}{T}$ approaches zero as $T$ increases since this bound is sublinear in $T$. Therefore, as long as we can prove the loss function is convex with respect to $\theta$, then we can add constraints to the variable norm $\|\theta\|_2$ and the gradient norm $\|\nabla_{\theta}\|$, and our algorithm will be no regret. Next, we will prove the convexity of our forward and reverse cross entropy losses.

\subsection{Trajectory Forecasting Loss}
First, we will consider the trajectory forecasting loss. To perform online learning for our model, we pretrain the trajectory simulator $f_\pi$ and fix its parameters, and only add a learnable linear layer to transform the $\mu_t$ output by the policy $\pi$, so $x_t$ can be rewritten as
\begin{equation}
    x_t = \hat\mu_t + \sigma_t z_t = x_{t-1} + \theta_\mu \mu_t + \sigma_t z_t \, .
\end{equation}
where $\theta_\mu$ is the learnable parameter. The policy output $\mu_t$ and $\sigma_t$ are fixed during online learning.

\paragraph{Trajectory Forward Cross Entropy}
For the forward cross entropy, we will not use the reparameterization trick and change-of-variable formula to analyze its convexity. Instead, it can be directly written as follows:
\begin{equation}
\begin{aligned}
    H(p, q_\pi) &= \mathbb{E}_{\tilde{x}\sim p} -\log q_\pi(\tilde{x}|\phi) \\
    &= \mathbb{E}_{\tilde{x}\sim p} \sum_{t=1}^T -\log N(\tilde{x}_t; \hat{\mu}_t, \Sigma=\sigma_t\sigma_t^T) \\
    &= \mathbb{E}_{\tilde{x}\sim p} \sum_{t=1}^T -\log \frac{\exp(-\frac{1}{2}(\tilde{x}_t-\hat{\mu}_t)^T\Sigma^{-1}(\tilde{x}_t-\hat{\mu}_t))}{\sqrt{(2\pi)^3 |\Sigma|}} \\
    &= \mathbb{E}_{\tilde{x}\sim p} \sum_{t=1}^T \frac{1}{2} (\log|\Sigma| + (\tilde{x}_t-\hat{\mu}_t)^T\Sigma^{-1}(\tilde{x}_t-\hat{\mu}_t) + 3\log2\pi) \, .
\end{aligned}
\end{equation}
Notice that since $\tilde{x}$ is sampled from $p$ rather than $q_\pi$, we can decompose the objective function over $t$, and we only need to demonstrate the convexity of the loss function at each time step. For convenience, we will omit the subscript $t$ in the following proof. 

In our setting, 
$\tilde{x}_t, \hat\mu_t \in \mathbb{R}^{3\times1}, \sigma_t \in \mathbb{R}^{3\times3}$. If we use a general linear layer, i.e., $\theta_\mu \in \mathbb{R}^{3\times3}$, the gradient of the objective is:
\begin{equation}
\begin{aligned}
    \nabla_{\theta_\mu} &= -\Sigma^{-1}(x_t - \hat{\mu}_t)\mu_t^T \\
    &= -\Sigma^{-1}(x_t - (x_{t-1} + \theta_\mu\mu_t))\mu_t^T \, .
\end{aligned}
\end{equation}
The Hessian matrix of the objective is
\begin{equation}
    \mathbf{H} = \nabla_{\theta_\mu}^2 = (\mu\mu^T \otimes \Sigma^{-1}) \, ,
\end{equation}
where $\otimes$ denotes the Kronecker product and $\mathbf{H} \in \mathbb{R}^{9\times9}$. We use $A$ to denote $\mu\mu^T$ and $B$ to denote $\Sigma^{-1}$. With the properties of the Kronecker product and the trace operator, we can prove the positive semidefiniteness of $\mathbf{H}$ as follows: for any $x \in \mathbb{R}^{3\times 3}$, we have

\begin{equation}
\begin{aligned}
    &\text{vec}(x^T)\cdot(A \otimes B)\cdot\text{vec}(x) \\
    =& \text{vec}(x^T)\cdot\text{vec}(AxB) \\
    =& \text{Tr}(x^TAxB) \\
    =& \text{Tr}(x^T\mu\mu^Tx\Sigma^{-1}) \\
    =& \text{Tr}(\mu^Tx\Sigma^{-1}x^T\mu) \\
    =& (x^T\mu)^T\Sigma^{-1}(x^T\mu)\\
    \geq& 0 \,,
\end{aligned}
\end{equation}
where we use the fact that since $\Sigma=\sigma_t\sigma_t^T$ is positive definite, its inverse $\Sigma^{-1}$ is also positive definite. Because the Hessian of the objective is positive semidefinite, our algorithm's forward cross entropy loss of trajectory forecasting is convex with regard to the linear layer's parameter $\theta_\mu$. 

\paragraph{Trajectory Reverse Cross Entropy}

The original reverse cross entropy of trajectory forecasting can be written as follows:
\begin{equation}
\begin{aligned}
    H(q_\pi, \Tilde{p}) &= -\mathbb{E}_{x\sim q_\pi}\log \Tilde{p}(x) \\
    &= -\mathbb{E}_{ x\sim q_\pi}\sum_{t=1}^T \log N(x_t; \mu = \tilde{x}_t, \Sigma = \lambda I) \, .
\end{aligned}
\end{equation}
Notice that $x_t$ is generated from $\mu_t$ and $\sigma_t$, which are functions of past generated positions $x_{t-1-H:t-1}$. So the reverse cross entropy contains complex nonlinear operations and it is hard to guarantee its convexity. To tackle this problem, at each time step $t$, we sample past trajectories $x_{1:t-1}$ from the true data distribution and only sample position $x_t$ from our policy, i.e. $x_{1:t-1} \sim p, x_t \sim q_\pi$. If we write down the reverse cross entropy under this setting, we will find it is $c_1(\mu_t+\sigma_t z_t - x_t)^T\Sigma^{-1}(\mu_t+\sigma_t z_t - x_t) + c_2$ ($c_1, c_2$ are constants), which only differs from the forward cross entropy $(x_t - \mu_t)^T\Sigma^{-1}(x_t - \mu_t)$ with a constant term. Thus, we can also prove the convexity of the reverse cross entropy using the convexity of the forward cross entropy.

\subsection{Action Forecasting Loss}
\paragraph{Action Forward Cross Entropy} Recall that the forward cross entropy loss of action forecasting is defined as

\begin{equation}
\begin{aligned}
    &H\left(p, q_\kappa\right) = -\mathbb{E}_{(x, a)\sim p}\log q\left(a|x, \phi\right) \\
    &= -\mathbb{E}_{(x, a) \sim p}\sum_{t,c} \log \tau\! \left(\sum_{i=1}^2\frac{u_{t,c, i}\left(\chi_t\right)}{a_{t,c,i}^\tau}\right)^{\!-\!2}\prod_{i=1}^2\left(\frac{u_{t,c, i}\left(\chi_t\right)}{a_{t,c,i}^{\tau+1}}\right),
\end{aligned}
\end{equation}
where $\kappa$ is our action policy (which maps context $\chi$ to action logits $u$), $a$ is the true action label, and $\tau$ is the Gumbel-Softmax temperature.

Similar to trajectory forecasting, we apply an affine transformation on $u$. Since $a$ is sampled from $p$ and there is no correlationship among actions if we apply the action-wise affine transformation, we can decompose the loss function over time and actions, and simply analyze the convexity of the loss for a single action class $c$ at a single time step $t$. Thus, we drop the subscripts $t,c$ and use $u_1, u_2$ to represent the action probabilities output by the policy $\kappa$. Since $u_1, u_2$ are generated by the softmax operation on the last layer's output $v_1, v_2$, we use parameters $\theta_1, \theta_2$ to transform the last layer before the softmax operation, and the new action probabilities are defined as
\begin{equation}
\begin{aligned}
    u_1 &= \frac{e^{\theta_1v_1}}{e^{\theta_1v_1} + e^{\theta_2v_2}} = \frac{w_1}{w_1 + w_2} \, ,\\
    u_2 &= \frac{e^{\theta_2v_2}}{e^{\theta_1v_1} + e^{\theta_2v_2}} = \frac{w_2}{w_1 + w_2} \, .
\end{aligned}
\end{equation}
Thus, the action forward cross entropy for a single action class $c$ at time step $t$ can be written as
\begin{equation}
\begin{aligned}
    H(\theta_1, \theta_2) &= -\log(c_1 u_1 + c_2 u_2)^{-2} (c_1'u_1)(c_2'u_2) \\
    &= -\log \frac{w_1w_2}{(c_1w_1 + c_2w_2)^2} \\
    &= -\log w_1w_2 + 2\log(c_1w_1+c_2w_2)\, ,
\end{aligned}
\end{equation}
where $c_1 = a_1^\tau, c_2 = a_2^\tau, c_1' = a_1^{\tau + 1}, c_2' = a_2^{\tau + 1}$.
Since $-\log w_1w_2 = -(\theta_1v_1 + \theta_2v_2)$ is a linear function, which is clearly convex, we only need to prove the convexity of $\log (c_1w_1 + c_2w_2)$ (sum is an operation preserving convexity).
The Hessian matrix of $\log (c_1w_1 + c_2w_2)$ is:
\begin{equation}
    \mathbf{H} = \begin{bmatrix}
    \frac{c_1c_2v_1^2A}{S^2} & \frac{-c_1c_2v_1v_2A}{S^2} \\
    \frac{-c_1c_2v_1v_2A}{S^2} & \frac{c_1c_2v_2^2A}{S^2} \\
    \end{bmatrix} \, ,
\end{equation}
where $A = e^{\theta_1v_1 + \theta_2v_2}, S = c_1e^{\theta_1v_1} + c_2e^{\theta_2v_2}$. We can prove its positive semidefiniteness by definition: for any $x \in \mathbb{R}^2$, we have
\begin{align*}
    x^T\mathbf{H}x &= \frac{c_1c_2A}{S^2}\begin{bmatrix}
    x_1 & x_2
    \end{bmatrix}^T
    \begin{bmatrix}
    v_1^2 & -v_1v_2\\
    -v_1v_2 & v_2^2
    \end{bmatrix}
    \begin{bmatrix}
    x_1 \\ x_2
    \end{bmatrix}\\
    &= \frac{c_1c_2A}{S^2}(v_1x_1 - v_2x_2)^2 \geq 0\,.
\end{align*}
Thus, our algorithm's forward cross entropy loss of action forecasting is also convex.

In summary, when we apply above linear transformations on the networks, our trajectory forward cross entropy, modified trajectory reverse cross entropy and action forward cross entropy are convex with respect to the parameters of the transformations. As a result, our model can perform online learning with the sum of these losses with theoretical guarantee of no-regret.

\clearpage
\section{Network Architecture Details}
\label{apsec:arch}
\begin{table}[ht!]
\centering
\ra{1.2}
\footnotesize
\begin{tabular}{@{}lllll@{}}\toprule
Component & Input[dimensionality] & Layer or Operation & Output[dimensionality] & Details \\ \midrule
\multicolumn{5}{l}{Trajectory Simulator} \\ \midrule
Traj & $[K, B, P, 3]$ & RNN & $[K, B, 100]$ & GRU cell, tanh activation \\
TrajFeat & $[K, B, 100]$ & FC & $[K, B, 200]$ & ReLU activation \\
TrajFeat & $[K, B, 200]$ & FC & $[K, B, 12]$ & Identity activation $\to \mu \in \mathbb{R}^{3}, s \in \mathbb{R}^{3\times3}$ \\ 
\multicolumn{5}{l}{\emph{Trajectory generation}: $x_t = x_{t-1} + \mu_t + \text{expm}(\text{softclip}(s_t)) \cdot z_t, \quad z_t \sim \mathcal{N}$} \\ \midrule
\multicolumn{5}{l}{Action Simulator} \\ \midrule
Image & $[B, 4, H, W, 3]$ & ResNet-50 \cite{he2016deep} & $[B, 4, 400]$ \\
ImageFeat & $[B, 4, 400]$ & FC & $[B, 400]$ & ReLU $\to$ ImageConsensus (a) \\
Traj & $[K, B, P, 3]$ & FC & $[K, B, 200]$ & ReLU \\
TrajEnc & $[K, B, 200]$ & FC & $[K, B, 200]$ & ReLU $\to$ TrajEncoding (b) \\
TrajFeat, ActFeat & a, b & Tile$(a)\oplus(b)$ & $[K, B, 600]$ & Concatenate($\oplus$) \\
JointFeat & $[K, B, 600]$ & FC & $[K, B, 500]$ & ReLU \\
JointFeat & $[K, B, 500]$ & FC & $[K, B, C_a \times 2]$ & Identity $\to v$\\
Action logits $[K, B, C_a \times 2]$ & Softmax & $[K, B, C_a \times 2]$ & Action Probability $u$ \\
\multicolumn{5}{l}{\emph{Action generation}: $a_{t, c} = \text{softmax} ((\log(u_{t, c}) + g_{t, c})/ \tau),\quad g_{t, c} \sim \mathcal{G}$} \\
\bottomrule
\end{tabular}
\vspace{3mm}
\caption{\textbf{Network architecture details.} Layers are arranged from top to bottom. We use the following hyper-parameters: sample number $K=12$, batch size $B=10$, past trajectory context horizon $P=10$, image size $H=W=224$, and the number of action classes $C_a=122$.}
\label{tab:arch_details}
\end{table}

\subsection{Trajectory Network}
Recall our trajectory simulator $f_\pi$:
\begin{equation*}
    x_t = \mu_t\left(\psi_t;\theta\right) + \sigma_t\left(\psi_t;\theta\right)z_t \,.
\end{equation*}
We assume that people tend to be still, so we model $\mu_t$ as: $\mu_t = x_{t-1} + \bar\mu_t$, where $\bar\mu_t$ can be interpreted as a \emph{velocity}. To ensure positive-definiteness of $\sigma_t$, we use the matrix exponential: $\sigma_t = \text{expm}(s_t + s_t^T)$. To enhance numerical stability, we soft-clip $s_t$ before calculating $\sigma_t$ with the following formula: $\text{softclip}(s, L) = \frac{s}{\log\sum\exp\left( \text{softmax}(1, \|s\|/L)\right)}$ (we use $L=5$) and also add a minimum precision identity matrix $\epsilon I$ to $\sigma_t$ before calculating the inverse of $\sigma_t$. 

We use a GRU to encode past positions  $x_{t-P:t-1} \in \mathbb{R}^{10\times3}$ with 100 hidden units and a 2-layer MLP to generate $\bar\mu_t \in \mathbb{R}^{3}$ and $\sigma_t \in \mathbb{R}^{3\times3}$ with 200 hidden units. The activation function is tanh for the GRU and ReLU for the MLP. The network architecture details can also be found in Table~\ref{tab:arch_details}.

\subsection{Action Network}

The ConvNet we use in our action network is ResNet-50 \cite{he2016deep} and the network architecture is mainly based on Temporal Segment Networks (TSN) \cite{wang2016temporal}. Past observed images of 2 seconds are sampled with 2 fps and cropped to $224 \times 224$. As a result, 4 past images in total are passed through the ConvNet separately. The output of the ConvNet is 400 dimensional for each image, and then a stacked 1600 dimensional feature is passed through a fully-connected layer to generate a 400 dimensional segmental consensus. 
Our action network also uses the past trajectory $x \in \mathbb{R}^{P\times3}$, which are passed through an MLP with a single 200-dim hidden layer with ReLU activation to output a 200-dim encoding. Finally, a 2-layer MLP takes the segmental consensus and the past trajetory encoding as input and outputs the action class scores. For this MLP, the hidden layer has 500 units and also use ReLU activation function. The output of the action network is $a \in \mathbb{R}^{C_a \times 2}$. In our setting, $P = 10, T = 5, C_a = 122$.
In one of the baseline, the trajectory-independent action forecasting model (\emph{Ours(S)}), the action network does not depend on trajectories, so the action network only takes the segmental consensus as input in that setting.

The CVAE baseline follows the same network structure as our model to encode context and decode context to generate positions and actions at each time step. The only difference is that the CVAE baseline uses the context encoding to generate the mean and standard deviation of latent variables with one fully connected layer. Following \cite{lee2017desire}, another fully connected layer with sigmoid activation is applied to map latent variables to a feature of the same dimensions as the context encoding. They are combined via element-wise multiplication.

\section{Other Details}

\subsection{Evaluation Metrics}
We calculate the example-based precision and recall in the same way as \cite{zhang2014review}:
$$\begin{aligned}
    \text{precision} &= \frac{1}{NT_a} \sum_{N,T_a} \frac{\Sigma_{C_a} tp}{\Sigma_{Ca} (tp + fp)}\,,\\ \text{recall} &= \frac{1}{NT_a} \sum_{N,T_a} \frac{\Sigma_{C_a} tp}{\Sigma_{C_a} (tp + fn)}\,,
\end{aligned}$$
where $N$ is the number of examples, and $T_a$ is the action forecasting horizon. $tp$, $fp$, $fn$ is the number of true positives, false positives, and false negatives respectively. 

\subsection{Data Augmentation}
Data augmentation can generate diverse training samples and alleviate over-fitting. In our implementation, we follow the same approach used in VGG \cite{simonyan2014very}: resize the original image to $256\times256$, randomly crop it to $224\times224$, and randomly flip it horizontally. Additionally, since we down-sample videos with a lower fps, we split the original sequence of images into 4 snippets with equal length, and randomly select images from each snippet during the training phase. This is similar to the approach used in TSN \cite{wang2016temporal}.

\subsection{Data Perturbation}
As mentioned in~\cite{rhinehart2018r2p2}, $H(p, q)$ is lower-bounded by $H(p)$, but it may be unbounded below since $H(p)$ may be arbitrarily negative. Thus, we also perturb the trajectories in training data with $-\mathbb{E}_{\eta\sim \mathcal{N}(0, \eta I)}\mathbb{E}_{x\sim p}\log q(x+\eta)$ and $\eta = 0.0001$. It eliminates singularity of the metric, because the perturbation distribution has a finite entropy. 

\subsection{Training Details}
In our prior distribution approximation, for trajectory reverse cross entropy $H(q_\pi, \tilde{p})$, we build $\Tilde{p}$ as a sequence of unimodal normal distributions with ground-truth trajectory $\tilde{x}$ as means, \emph{i.e.}, $\tilde{p}(x|\phi) = N(\cdot|\tilde{x}; \sigma I)$. We choose $\sigma = 0.01$. For action $H(q_\kappa, p)$, we also view each action happening at $t$ as a unimodal normal distribution in the time dimension. We choose the 0.5 as the scale factor. For the model trained with both forward and reverse cross entropy losses, we use $\beta=0.02$ for the trajectory reverse cross entropy and $\beta=0.1$ for the action reverse cross entropy.

We use Adam \cite{kingma2014adam} as the optimizer to perform gradient descent. The learning rate is set to 1e-4. The gradient is cut off from the action network to the trajectory network. When training the joint model, we use ResNet pretrained on ImageNet to initialize our ConvNet, and we also pretrain our trajectory network with trajectory forecasting. The number of epochs is 50 for training trajectory our joint model and 300 for training the separate model. The batch size is 16 and the sample number of each data is 12.  We report the results on the test set using the model that performs the best on the validation set.

\clearpage
\section{Additional Evaluation Results}

\begin{figure}[ht!]
\vspace{-3mm}
\centering
\includegraphics[width=0.9\linewidth]{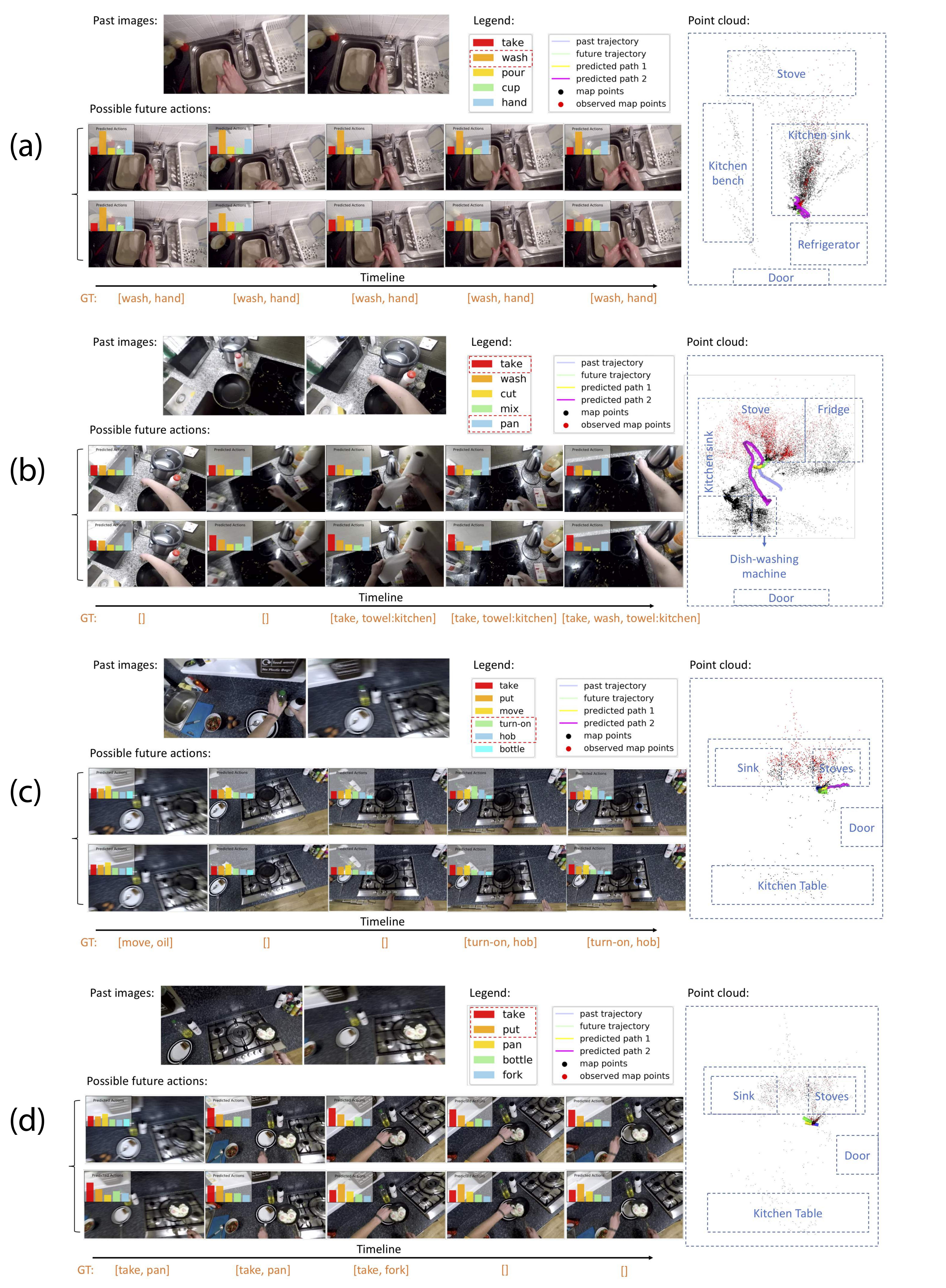}
\caption{\textbf{Forecasting results visualization.} The figure shows the additional visualization results of four examples. It shows how the forecasted trajectory influences the action distribution. In each example, the left top shows the observed images in the past 2 seconds, the left bottom shows the action distributions corresponding to two forecasted sample trajectories, and the right shows the point cloud of the scene and the forecasted trajectories.}
\vspace{-15mm}
\label{fig:vis_supp}
\end{figure}

\subsection{Diversity Analysis}
Our model can generate more diverse samples than the CVAE and the discriminative model, as measured by two new metrics
of diversity:
\begin{itemize}
    \item The number of similar action sequences (\emph{N.Act}). Two time-related (TR) action sequences count as similar if action $a$ occurs at time $t$ in both sequences. Two time-unrelated (TU) action sequences count as similar if action $a$ occurs at any time in both sequences. Similar to \cite{kingma2018glow}, we apply temperature scaling on the Gumbel noise $g$ with a factor of 0.3.
    \item The cosine similarity of generated samples (\emph{CoSim}). We use this metric to evaluate both actions and trajectories. We report the mean cosine similarity between ${12 \choose 2}$ pairs. We also report another number in parentheses by setting a similarity threshold $0.3$ to determine whether two samples are different and count the number of different samples.
\end{itemize}

\begin{table}[ht!]
\centering
\resizebox{.75\textwidth}{!}{
\centering
\begin{tabular}{@{}lccccccc@{}}\toprule
 & Trajectory && \multicolumn{4}{c}{Action} \\
 \cmidrule{2-2} \cmidrule{4-7}
  Method      & Traj CoSim ($\downarrow$) && N.Act (TR) ($\uparrow$) & N.Act (TU) ($\uparrow$) & Act CoSim (TR) ($\downarrow$) & Act CoSim (TU) ($\downarrow$)\\ \midrule
MRMC & --           && 0.548 & 1.236 & -- & -- \\
CVAE & 0.232 (2.83) && 0.772 & 1.419 & 0.618 (2.05) & 0.634 (2.39) \\
Ours & {\bf 0.168 (4.13)} && {\bf 1.651} & {\bf 7.463} &{\bf  0.291 (4.38)} & {\bf 0.115 (8.83)}  \\
\bottomrule
\end{tabular}
}
\vspace{1em}
\caption{\textbf{Sample diversity evaluation results.} $(\downarrow)$/($\uparrow$) denotes a metric for which lower/higher scores are better.}
\label{tab:diversity_results}
\end{table}

\subsection{Visualization Results}

Fig.~\ref{fig:vis_supp} shows additional visualization results of four examples. For each example, we show two sampled trajectories and their corresponding action distribution. In all these examples, the forecasted trajectory influences the action distribution in a meaningful way. 
In the first example (a), the person is washing hands. The \emph{wash} action will be  more likely to happen if the person stays still, just as indicated by the first sampled trajectory and its action distribution. If the person moves a lot such as in the second sampled trajectory, the model predicts the wash action is less likely to happen.
In the second example (b), the person moves less in the first forecasted trajectory than the second one. Thus, in the first trajectory, the object the person interacts with in the observed frames, i.e. \emph{pan}, will have a high probability in the future. In the second trajectory, although the model does not predict the ground-truth object \emph{towel:kitchen} correctly, the probability of \emph{take} increases and the probability of \emph{pan} decreases, which indicates the person may take something else because the person moves more in this forecasted trajectory.
In the third example (c), the probability of \emph{turn-on} and \emph{hob} increases when the person moves and then stops, as shown in the first sampled trajectory. However, if the person keeps moving like in the second trajectory, these two actions will both have a low probability.
In the fourth example (d), the large range of movement like the first sampled trajectory will lead to a higher probability of \emph{take} than \emph{put}, because when taking something the person has to move around to fetch the item.

\end{document}